\documentclass{article}

\usepackage[preprint]{neurips_2026}

\usepackage{amsmath}
\usepackage{amsthm}
\newtheorem{theorem}{Theorem}
\newtheorem{lemma}{Lemma}

\newtheorem{corollary}{Corollary}
\newtheorem{remark}{Remark}
\usepackage{graphicx}
\usepackage{bbm}
\usepackage[ruled,vlined]{algorithm2e}

\SetCommentSty{algcommentfont}

\usepackage[utf8]{inputenc} 
\usepackage[T1]{fontenc}    
\usepackage{hyperref}       
\usepackage{url}            
\usepackage{booktabs}       
\usepackage{amsfonts}       
\usepackage{nicefrac}       
\usepackage{microtype}      
\usepackage{xcolor}         

\usepackage{wrapfig}
\usepackage[table]{xcolor}
\usepackage{makecell}

\definecolor{mylightblue}{RGB}{191,205,218}
\definecolor{iqnblue}{RGB}{120,170,215}

\title{Quantile Geometry Regularization for Distributional Reinforcement Learning}

\author{%
  Zhaofan Zhang, Minghao Yang, Rufeng Chen, Sihong Xie, Hui Xiong \\
  Information Hub, AI Thrust \\
  Hong Kong University of Science and Technology (Guangzhou) \\
  Guangzhou, Guangdong, China
}

\begin{document}

\maketitle

\begin{abstract}
  Quantile-based distributional reinforcement learning methods learn return distributions through sampled quantile regression, but their bootstrapped target quantiles may induce distorted or degenerate distribution estimates. We propose Robust Quantile-based Implicit Quantile Networks (RQIQN), a lightweight Wasserstein distributionally robust enhancement boosted from a quantile estimation perspective. We first reinterpret a snapshot of IQN loss as a collection of local empirical quantile estimation problems over sampled current fractions. We then robustify each local slot with a Wasserstein distributionally robust quantile estimation formulation, yielding a closed-form, fraction-dependent correction to the Bellman target. 
  This correction directly addresses distributional degeneration: its median-antisymmetry preserves the risk-neutral quantile average, while its monotonicity enlarges upper--lower quantile gaps and counteracts collapsed distributional spread. RQIQN thus regularizes quantile geometry without changing the underlying value objective or requiring additional sample-set reconstruction. Finally, we empirically show that the proposed RQIQN outperforms other existing quantile-based distributional reinforcement learning algorithms in risk-sensitive navigation and Atari games.
\end{abstract}

\section{Introduction}

\begin{figure}[t]
\centering
\includegraphics[width=0.98\linewidth]{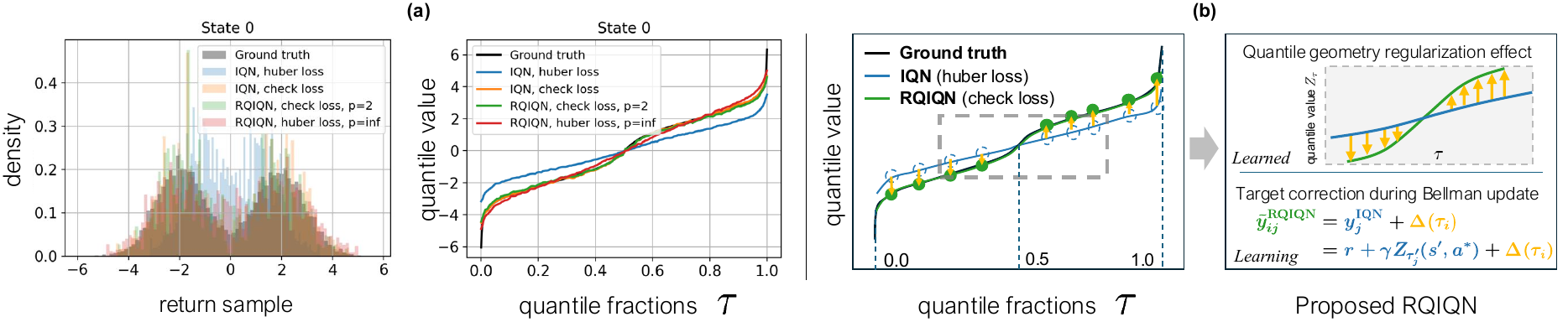}
\vspace{-0.10cm}
\caption{An illustration of \textbf{(a)} distribution degeneration at state $0$ and
\textbf{(b)} how the proposed RQIQN correction modulates quantile geometry. The samples visualization of fitted return distributions are from a four-state chain MDP with deterministic transitions under a unique action. State transitions are directional and sequential, progressing from state \(0\) to state \(3\). 
Rewards are zero except at the terminal state, where
$r \sim \frac{1}{2}\mathcal{N}(-2,1)+\frac{1}{2}\mathcal{N}(2,1)$. In \textbf{(b)}, each agent uses its default training loss.
}
\label{fig1}
\end{figure}

Distributional reinforcement learning (DistRL) models the full distribution of
discounted returns rather than only its expectation, providing a richer basis
for learning under stochasticity and risk. 
A useful way to understand practical DistRL algorithms is through the distinction
between return distributions and statistics of those distributions.
The distributional Bellman operator acts on full
return distributions, whereas scalable algorithms usually propagate
finite-dimensional summaries, such as atoms, quantiles, or expectiles. 
Existing DistRL algorithms differ
mainly in how they represent and update return distributions, including
categorical atoms, quantile functions, expectiles, and sample-set
representations. 
Among these choices, quantile-based methods are particularly
appealing because quantile functions naturally align with Wasserstein geometry,
do not require a fixed value support, and provide direct access to different
regions of the return distribution. 
Within this scope, Implicit Quantile Networks (IQN)~\cite{dabney2018implicit} further extends QR-DQN~\cite{dabney2018distributional} with fixed-grid quantiles by learning a continuous quantile function through sampled quantile fractions $\tau\sim U([0,1])$.
Generally, a set of
statistics is Bellman-closed if the statistics of the Bellman-updated
distribution can be computed solely from the same statistics of the next-state
distribution and the reward, without reconstructing the full distribution.
This nature is crucial since it allows recursive dynamic
programming (Eq.~\eqref{zdef}) in the chosen statistic space. 
However, finite quantile statistics are not Bellman-closed~\cite{rowland2019statistics}, which can lead to distribution degeneration and biased estimates of distributional spread (Figure~\ref{fig1}(a)).

In quantile-based DistRL, deep Q-learning~\cite{mnih2013playing} is extended from scalar value regression to distributional quantile fitting. 
For each transition, the agent forms pairwise TD errors between current quantile estimates and bootstrapped target quantiles, and minimizes a quantile regression loss (Eq.~\eqref{eq:IQNloss}). 
Thus, the TD update has an explicit quantile-estimation interpretation, with each sampled fraction corresponding to a local quantile fitting problem.
In this work, we revisit IQN through the lens of local quantile estimation. 
Specifically, the IQN loss (Eq.\eqref{eq:IQNloss}) can be interpreted exactly at the empirical-loss level as a collection of empirical quantile-estimation slots, one for each sampled current quantile fraction $\tau_i$.
Each slot fits the $\tau_i$ located quantile of the bootstrapped Bellman target induced by target quantile fractions $\{\tau'_j\}_{j=1}^{N'}$. 
This view reveals that IQN performs local quantile estimation over a model-generated empirical target distribution formed by finitely many bootstrapped target quantile values. From this perspective, distribution degeneration can be understood as accumulated bias in the learned quantile values, which distorts the geometry of the represented return distribution.

We propose RQIQN (Figure~\ref{fig1}(b)) to address the distribution degeneration problem, namely biased quantile representation, by robustifying each local quantile fitting problem against worst-case perturbations of the empirical Bellman target distribution within a Wasserstein ambiguity set.
The resulting closed-form quantile correction regularizes the learned return distribution by stabilizing finite-sample quantile estimates.
Instead of introducing computationally intensive schemes by modifying the distributional Bellman operator or distributional representation in existing work~\cite{rowland2019statistics,jullien2023distributional,nguyen2020distributional}, RQIQN replaces each local quantile regression slot with a Wasserstein distributionally robust quantile estimation problem around the empirical Bellman target law. For the check-loss formulation\footnote{Namely, quantile loss. It's different from quantile Huber loss (Huberized quantile regression loss).}, this robust slot admits a closed-form fraction-dependent correction $\Delta_p(\tau;\epsilon)$, where $p$ is the Wasserstein order, and $\epsilon$ is the robustness radius that decays over time. Geometrically, the correction is median-antisymmetric and monotone in $\tau$. Hence it preserves the uniform quantile average while expanding upper--lower quantile gaps, providing a mean-neutral regularization of distributional spread.

\subsection{Related Work}

DistRL has developed rapidly as a powerful alternative to expectation-based value learning, with representative approaches modeling return distributions using categorical atoms, quantiles, expectiles, or sample-based representations~\cite{bellemare2017distributional,
dabney2018distributional,dabney2018implicit,rowland2019statistics,yang2019fully}. Among them,
IQN~\cite{dabney2018implicit} has become particularly influential due
to its flexible implicit quantile-function representation, which supports arbitrary quantile sampling
and naturally enables risk-sensitive policies through distortion risk measures. 
IQN has been widely applied in tasks that require risk-sensitive decision-making, such as autonomous surface navigation~\cite{lin2023robust,zhang2025perturbation} and quadrupedal locomotion~\cite{shi2024robust}.
However, IQN still inherits a limitation of quantile-based DistRL: finite
quantile statistics are not Bellman-closed. 

Several lines of work have addressed the difficulty caused by non-Bellman-closed
distributional statistics by either changing the learned statistics or introducing
an explicit statistics-to-samples interface. Expectiles, introduced as asymmetric
least-squares location statistics~\cite{newey1987asymmetric}, are attractive
because they provide smoother $L_2$-based fitting than check-loss quantile
regression, and their symmetric case $\tau=1/2$ coincides with the mean. This property
makes expectiles, useful for stable value estimation, since the mean is the statistic
used for risk-neutral action selection. However, a finite set of expectiles is
still not Bellman-closed in general, and expectile values cannot be directly
interpreted as samples from the return distribution. To make expectiles usable
for Bellman backups, ER-DQN~\cite{rowland2019statistics} recover Bellman
target samples through an imputation step, which requires solving costly
nonlinear systems.
IEQN~\cite{jullien2023distributional} instead jointly learns expectiles and
quantiles to retain efficient $L_2$-based learning while avoiding explicit
imputation, at the cost of additional prediction heads and auxiliary coupling
losses. Sample-set methods, such as MMD~\cite{nguyen2020distributional} and
MWG~\cite{zhang2024distributional}, take a different route by operating on
explicit return samples, thereby bypassing the need to propagate non-closed
statistics as samples. However, deterministic sample sets may limit stochastic
target diversity, while Gaussian-mixture variants rely on EM-style
projection~\cite{dempster1977maximum} and sample augmentation, introducing
additional computational and algorithmic complexity.

\section{Problem Setup and Main Analysis}
\label{gen_inst}

\subsection{Distributional Reinforcement Learning}

We consider a Markov decision process (MDP)
\(
\mathcal{M}=(\mathcal{S},\mathcal{A},\mathcal{P},\mathbb{R},\gamma)
\),
where \(\mathcal{S}\) and \(\mathcal{A}\) denote the state and action spaces,
\(\mathcal{P}(\cdot\mid s,a)\) is the transition kernel,
\(\mathbb{R}(\cdot\mid s,a)\) denotes the reward distribution, and
\(\gamma\in[0,1)\) is the discount factor. For a policy \(\pi\), the
state-action return is the random variable
\(
    Z^\pi(s,a)
    \overset{D}{=}
    \sum_{t=0}^{\infty}\gamma^t R_t
\),
\(S_0=s\), \(A_0=a\),
where \(
    \overset{D}{=}
\) denotes equality in distribution and the transition process can be explained by \(S_{t+1}\sim\mathcal{P}(\cdot\mid S_t,A_t)\),
\(R_t\sim\mathbb{R}(\cdot\mid S_t,A_t)\),
\(A_t\sim\pi(\cdot\mid S_t)\) for \(t\geq 1\).
Standard reinforcement
learning typically optimizes only the first moment of this return
distribution. In particular, the action-value function is defined as
\(
    Q^\pi(s,a)=\mathbb{E}_Z\!\left[Z^\pi(s,a)\right].
\)
The optimal action-value function satisfies the Bellman optimality equation
\(
    Q(s,a)
    =
    \mathbb E_{R\sim\mathbb R(\cdot\mid s,a),\,
S'\sim\mathcal P(\cdot\mid s,a)}\!\left[
        R+\gamma\max_{a'\in\mathcal{A}}Q(S',a')
        \,\middle|\, S=s,A=a
    \right]
\). This expectation-based formulation
compresses the full return distribution into a single scalar,
discarding distributional information induced by stochastic transitions and rewards.

DistRL extends standard RL by modeling the \emph{return distribution} \(Z\) instead of only its expectation. Formally, \(Z\) can be updated by dynamic programming, with distributional Bellman optimality operator\footnote{Uppercase letters $S$ and $A$ denote random variables, and lowercase letters denote their realized values.} defined by
\begin{equation}
\label{zdef}
    \mathcal{T}^\pi Z(s,a) \stackrel{D}{=} R(s,a) + \gamma Z(S', A').
\end{equation}
We can compute the optimal return distribution by using the distributional Bellman optimality operator \(\mathcal{T}\) defined as
\begin{equation}
\label{operator}
    \mathcal{T}Z(s,a) \stackrel{D}{=} R(s,a) + \gamma Z\left( S', a^* \right),\qquad a^*=\operatorname*{argmax}_{a'} \mathbb{E}_Z[Z(S',a')].
\end{equation}
As a representative quantile-based DistRL method, Implicit Quantile Networks (IQN)~\cite{dabney2018implicit} provides a flexible distributional representation through the quantile function of the return distribution. 
Let $F^{-1}_{Z(s,a)}(\tau)$ denote the quantile function of the random return $Z(s,a)$ at quantile fraction $\tau$. For notational simplicity, we write
\(
Z_{\tau}(s,a) := F^{-1}_{Z(s,a)}(\tau)
\),
so that when $\tau \sim U([0,1])$, the sampled quantile value $Z_{\tau}(s,a)$ follows the return distribution $Z(s,a)$. Based on this implicit quantile representation, IQN parameterizes this quantile function with a
neural network $Z_\tau(x,a;\theta)\approx Z_{\tau}(s,a)$. For brevity, we use $Z_{\tau}(s,a)$ directly.
Since IQN samples quantile fractions explicitly, risk-sensitive behavior can be induced by modifying the quantile sampling range. 
A common choice is the conditional value-at-risk (CVaR) distortion, which maps a uniformly sampled fraction $\tau\sim U([0,1])$ to
\(
    \tilde{\tau} = \eta \tau
\),
thereby restricting sampling to the lower-tail quantile region $U([0,\eta])$ for risk-averse control. In this work, unless explicitly stated, fractions are sampled without distortion.

Throughout this work, we consider two quantile regression losses related to DistRL. The first is
the check loss,
\[
    \rho_\tau(u)
    =
    u\left(\tau-\mathbbm{1}_{\{u<0\}}\right)
    =
    \left|\tau-\mathbbm{1}_{\{u<0\}}\right|\,|u|,
\]
which gives the exact quantile-regression objective. The second is the standard
quantile Huber~\cite{huber1992robust} loss used in practical QR-DQN/IQN implementations:
\[
    \rho_\tau^\kappa(u)
    =
    \left|\tau-\mathbbm{1}_{\{u<0\}}\right|
    \frac{\mathcal H_\kappa(u)}{\kappa},
    \qquad
    \mathcal H_\kappa(u)
    =
    \begin{cases}
    \frac12 u^2, & |u|\le \kappa,\\[1mm]
    \kappa\left(|u|-\frac12\kappa\right), & |u|>\kappa.
    \end{cases}
\]
Here, $\kappa>0$ is the
Huber threshold. The check loss preserves the exact quantile-estimation
interpretation, while the Huberized loss provides a smoother surrogate for
numerical stability and obtaining higher return.

\subsection{Distribution Degeneration with Quantile-based Representation}\label{sec:degeneration}

In quantile-based DistRL, the distributional Bellman update is implemented
through TD target fitting, where current quantile estimates are regressed toward
bootstrapped target quantiles. However, finite quantile sets are not
Bellman-closed in general, and quantiles are statistics rather than samples. Thus, unless the Bellman-updated distribution is
explicitly projected back to the quantile representation or reconstructed through
an imputed target distribution, approximation bias may accumulate and distort
the learned return distribution~\cite{rowland2019statistics}.
In practice, this issue is exacerbated by the
Huberized quantile regression loss, which no longer preserves the exact guarantees of quantile regression. 
The learned return distribution may consequently collapse toward its mean, discarding tail information.

The degeneration risk of return distribution motivates a robustness-oriented view of quantile-based
distributional learning, in which mitigating degeneration amounts to stabilizing the quantile estimates produced by the proceeding of Bellman updates. 
In Figure~\ref{fig1}, IQN (\textcolor{iqnblue}{depicted in blue}) suffers from worse variance degradation, whereas the RQIQN under two loss settings demonstrates robustness in terms of distribution geometry. The detailed formulation of RQIQN is provided in Section~\ref{RQIQNalgorithm}.




\subsection{The Quantile Estimation Inspired Loss}

Given a transition $(s,a,r,s')$, IQN samples current fractions
$\{\tau_i\}_{i=1}^{N}$ for the quantile levels to be fitted at $(s,a)$ and target
fractions $\{\tau'_j\}_{j=1}^{N'}$ for bootstrapped next-state quantile targets,
both from $U([0,1])$, and defines the IQN loss with TD error \(
    \delta_{ij}
\) as
\begin{equation}\label{eq:IQNloss}
\mathcal{L}_{\mathrm{IQN}}(s,a,r,s')
=
\frac{1}{N'}
\sum_{i=1}^{N}
\sum_{j=1}^{N'}
\rho^{\kappa}_{\tau_i}
\bigl(
\delta_{ij}
\bigr)
=
\sum_{i=1}^{N} \mathbb{E}_{\tau'}\left[\rho_{\tau_i}^{\kappa}\left(\delta_{ij}\right)\right], 
    \quad
    \delta_{ij}
=
r + \gamma Z_{\tau'_j}(s',a') - Z_{\tau_i}(s,a).
\end{equation}
Similar to the TD loss in deep Q-learning~\cite{mnih2013playing}, it is mentioned in QR-DQN~\cite{dabney2018distributional} that a nature in the loss function of quantile-based DistRL is to employ quantile
regression temporal difference learning (QRTD).
The update is computed for all pairs of $(Z_{\tau'_j}(s', a'), Z_{\tau_i}(s, a))$. At its essence, quantile estimation seeks the value corresponding to a given quantile level $\tau$ from samples drawn from a distribution. With respect to the TD error in Eq.~\eqref{eq:IQNloss}, if the quantile fraction $\tau_i$ is fixed, the update target is regarded as a sample, then the update can be interpreted as estimating the quantile at level $\tau_i$. Therefore, the overall IQN loss performs quantile estimation over all sampled quantile levels during the distributional Bellman update, which in turn yields a representation of the return distribution.

\begin{lemma}[Snapshot of Implicit Quantile Network loss as quantile estimation]
\label{lem:iqn_snapshot_local_qr}
For any transition \((s,a,r,s')\), consider a snapshot \(\mathbb{E}_{\tau'}\left[\rho_{\tau_i}^{\kappa}\left(\delta_{ij}\right)\right]\) of IQN loss in Eq.~\eqref{eq:IQNloss} with quantile fraction samples 
\(\tau'_j\sim U([0,1])\) and a fixed one \(\tau_i\in(0,1)\):
\[
    \delta_{ij}
    =
    y_j
    -
    q_i,
    \qquad
    y_j
    :=
    r + \gamma Z_{\tau'_j}(s', a'),
    \qquad
    q_i
    :=
    Z_{\tau_i}(s, a).
    \label{eq:td_snapshot}
\]

Let the empirical Bellman target be
\(
    \widehat{\mu}_{s,a}
    =
    \frac{1}{N'}
    \sum_{j=1}^{N'}
    \delta_{y_j}
    \label{eq:empirical_bellman_target_law}
\),
where \(\delta_{y_j}\) denotes a Dirac at \(y_j\). The empirical quantile regression slot of Eq.~\eqref{eq:IQNloss} at \(\tau_i\) is
\begin{equation}\label{eq:qr_slot}
    q_{\tau_i}^{0}
    \in
    \arg\min_{q\in\mathbb{R}}
    \mathbb{E}_{Y\sim\widehat{\mu}_{s,a}}
    \left[
        \rho_{\tau_i}(Y-q)
    \right],
\end{equation}
where $\rho_\tau(u)=u(\tau-\mathbbm{1}_{\{u<0\}})$
and any
minimizer \(q_{\tau_i}^{0}\) satisfies the empirical coverage condition
\[
    \mathbb{P}_{Y\sim \hat{\mu}_{s,a}}(Y<q_{\tau_i}^{0})
    \le
    \tau_i
    \le
    \mathbb{P}_{Y\sim \hat{\mu}_{s,a}}(Y\le q_{\tau_i}^{0}).
    \label{eq:local_qr_coverage}
\]

\end{lemma}
Summarized by Lemma~\ref{lem:iqn_snapshot_local_qr}, the IQN loss minimization
can be interpreted as simultaneously solving a collection of local
quantile regression problems over sampled fractions
\(\{\tau_i\}_{i=1}^{N}\). 


\subsection{Robustness of Quantile Geometry Regularization}\label{sec:qgr_robustness}

Building on the degeneration analysis in Section~\ref{sec:degeneration}, we
reinterpret the robustness issue of the return distribution as a sequence of local
robust numerical quantile estimation problems in Lemma~\ref{lem:iqn_snapshot_local_qr}. 
Specifically, once the
Bellman target distribution is fixed, each sampled quantile level
\(\tau\in(0,1)\) asks for a scalar estimate \(q_\tau\) that locally
represents the \(\tau\)-quantile of the target return law. 

Robust quantile estimation~\cite{tzavidis2010robust,john2015robustness,galarza2017robust} has been extensively studied in statistics literature, and is closely related to the robustness of IQN loss at a specific snapshot.
More recently, Wasserstein distributionally robust
quantile regression (WDRQR)~\cite{zhang2026wasserstein} shows
that, under a type-\(p\) Wasserstein ambiguity set, distributional
robustness induces an exact regularized reformulation for the
check-loss quantile regression objective. More importantly, this
robustness effect is not merely a global penalty. The
robust solution differs from the corresponding regularized nominal
solution through a radius-dependent location adjustment, while the
check loss is essentially the unique convex loss class that admits
such an additive Wasserstein regularization phenomenon.
Next, we present Theorem~\ref{thm:wdrqr} in a concise form based on WDRQR, making its connection to the statistical intuition more explicit.

\begin{theorem}[Wasserstein-robust local quantile correction]
\label{thm:wdrqr}

For $\epsilon \ge 0$, $p \in (1,\infty]$, consider the Wasserstein distributionally robust quantile estimation slot at \(\tau_i\) from Eq.~\eqref{eq:qr_slot}
\begin{equation}\label{eq:wrqr_slot}
q_{\tau_i}^{\epsilon}
\in
\arg\min_{q\in\mathbb{R}}
\sup_{\nu \in B_p(\hat{\mu}_{s,a},\epsilon)}
\mathbb{E}_{Y\sim\nu}\!\left[\rho_{\tau_i}(Y-q)\right],
\end{equation}
where $B_p(\hat{\mu}_{s,a},\epsilon)$ denotes the type-$p$ Wasserstein ambiguity set
\(
B_p(\hat{\mu}_{s,a},\epsilon)
:=
\{\nu \in \mathcal{P}(\mathbb{R}) : W_p(\nu,\hat{\mu}_{s,a}) \le \epsilon\}.
\)
For each sampled fraction $\tau_i$, any robust minimizer admits the explicit location correction
\begin{equation}\label{eq:delta}
q_{\tau_i}^{\epsilon}
=
q_{\tau_i}^0 + \Delta_p(\tau_i,\epsilon),
\end{equation}
where the robust term $\Delta_p$ can be expressed as
\begin{equation}\label{eq:robustterm}
\Delta_p(\tau,\epsilon)
=
\frac{\epsilon}{q}
\big(\tau^q-(1-\tau)^q\big)\,
c_{\tau,p}^{\,1-q},
\qquad
\frac{1}{p}+\frac{1}{q}=1,
\end{equation} \\
and
\begin{equation}
c_{\tau,p}
=
\begin{cases}\label{eq:needdecay}
\big(\tau^q(1-\tau)+\tau(1-\tau)^q\big)^{1/q},
& p\in(1,\infty),\\[4pt]
2\tau(1-\tau),
& p=\infty.
\end{cases}
\end{equation}
\end{theorem}
For each sampled fraction $\tau_i$, the Wasserstein distributionally robust quantile estimation problem in Theorem~\ref{thm:wdrqr} can be written as
\(
q_{\tau_i}^{\epsilon}
\in
\arg\min_{q\in\mathbb{R}}
\frac{1}{N'}\sum_{j=1}^{N'}
\rho_{\tau_i}\!\left(y_j+\Delta_p(\tau_i,\epsilon)-q\right).
\)

Standard IQN estimates this quantity under the empirical target
induced by transition samples and corresponding target. In contrast, our goal
is to estimate a quantile that remains stable under local distributional
perturbations of the Bellman target. Thus, instead of directly
robustifying the infinite-dimensional return distribution, we robustify
the scalar quantile estimator that represents each local slice of the
learned quantile function.

\begin{remark}[Mean-neutral geometry modulation]
\label{rem:mean_neutral_geometry}
The robust correction term in Eq.~\eqref{eq:robustterm} is antisymmetric around the median:
\(
    \Delta_p(1-\tau;\epsilon)=-\Delta_p(\tau;\epsilon),
    \Delta_p(1/2;\epsilon)=0.
\)
Hence, under uniform quantile sampling,
\(
    \mathbb E_{\tau\sim\mathcal U(0,1)}
    [\Delta_p(\tau;\epsilon)]
    =
    \int_0^1 \Delta_p(\tau;\epsilon)d\tau
    =
    0.
\)
Therefore, the robust correction preserves the risk-neutral quantile average
\(
    \mathbb E_{\tau}
    \left[
        Z_\tau(s,a)+\Delta_p(\tau;\epsilon)
    \right]
    =
    \mathbb E_{\tau}
    \left[
        Z_\tau(s,a)
    \right].
\)
Moreover, since $\Delta_p(\tau;\epsilon)$ is nondecreasing in $\tau$, for
any $0<\tau_l<\tau_h<1$,
\(
\left[ Z_{\tau_h}(s,a)+\Delta_p(\tau_h;\epsilon)\right]
-\left[ Z_{\tau_l}(s,a)+\Delta_p(\tau_l;\epsilon)\right]
= Z_{\tau_h}(s,a)-Z_{\tau_l}(s,a)
+\Delta_p(\tau_h;\epsilon)-\Delta_p(\tau_l;\epsilon)
\),
with non-negative item $\Delta_p(\tau_h;\epsilon)-\Delta_p(\tau_l;\epsilon)\ge0$.
Thus, the robust correction expands upper--lower quantile gaps while leaving
the quantile average unchanged in expectation.
\end{remark}

\section{Algorithm}\label{RQIQNalgorithm}

In this section, we propose the practical algorithm for \textbf{RQIQN}. It can be readily adapted to other variants of QRTD based DistRL.

\begin{corollary}[The RQIQN Loss]\label{cor:RQIQNloss}
Following the form of Eq.~\eqref{eq:delta}, robust quantile-regression-based loss function for RQIQN is given by
\begin{equation}
\mathcal{L}_{\mathrm{RQIQN}}(s, a, r, s')
=
\frac{1}{N'}
\sum_{i=1}^{N}\sum_{j=1}^{N'}
\rho_{\tau_i}
\!\left(
r+\gamma Z_{\tau'_j}(s',a^*)
+\Delta_p(\tau_i,\epsilon)
-
Z_{\tau_i}(s,a)
\right).
\end{equation}

\end{corollary}

A key issue arises when correction-related term \(c_{\tau,p}\) in Eq.~\eqref{eq:needdecay}
is applied to continuously sampled quantile fractions in RQIQN. For
$p\in(1,\infty)$, the closed-form correction may become unbounded as
$\tau\rightarrow 0$ or $1$. This is undesirable in deep
distributional Q-learning, where quantile fractions are sampled from a continuous
space and bootstrapped targets are further affected by non-stationary network
updates. Moreover, since Q-learning itself is prone to overestimation, directly
amplifying extreme target quantiles may further destabilize TD
learning.

As discussed in Remark~\ref{rem:mean_neutral_geometry}, the robust correction provides
stronger correction intensity near the distributional tails, where quantiles correspond to
extreme return outcomes and are more sensitive to finite-sample uncertainty. In
our implementation, we focus on two representative cases, $p=2$ and
$p=\infty$. The case $p=\infty$ yields the bounded correction
$\Delta_\infty(\tau;\epsilon)=\epsilon(2\tau-1)$. For $p=2$, the raw correction
contains a denominator that vanishes near the endpoints. We therefore use a
bounded adaptation by reversing the endpoint behavior of the denominator:
\[
    \Delta_2(\tau;\epsilon)
    =
    \frac{\epsilon}{2}
    \frac{1-2\tau}
    {\sqrt{\tau^2+(1-\tau)^2}} .
\]
This adapted term remains finite for all $\tau\in[0,1]$, while preserving the
median-antisymmetry and monotonicity required by the geometric interpretation in
Remark~\ref{rem:mean_neutral_geometry}. To avoid directly increasing bootstrapped target values, \(\Delta_2(\tau;\epsilon)\) provides a subtractive prediction-side correction in the TD
residual. Under this sign convention, the effective corrected quantile geometry
still satisfies the same mean-neutral spread-modulation property.

Statistically, WDRQR~\cite{zhang2026wasserstein} suggests a power-law shrinkage of the Wasserstein radius with the empirical sample size, i.e., $\epsilon_N=O(N^{-1/2})$, deep distributional Q-learning induces non-stationary empirical targets through replay sampling, bootstrapping, and network updates. We therefore decouple $\epsilon$ from the nominal batch size and use a decay schedule, preserving strong early-stage robustness and gradually annealing the correction as training stabilizes.
Specifically, to implement adaptive robustness during training rather than using a fixed radius $\epsilon$, we adopt a time-dependent reverse-logistic decay schedule 
\[
\epsilon_t
=
\frac{\epsilon_0}{1+\exp(k(t-t_0))},
\]
where $t$ is the training step, $\epsilon_0$ denotes the initial robustness scale, $k>0$ controls the decay sharpness, and $t_0$ specifies the midpoint of the reverse-logistic transition.





\begin{algorithm}[t]
\caption{Robust Quantile Implicit Quantile Network Loss}
\label{alg:rqiqn_loss}
\DontPrintSemicolon

\KwIn{$(s,a,r,s')$, $\gamma\in[0,1)$, $N,N',K$, timestep $t > 0$, $\epsilon_t\ge0$, 
$p\in(1,\infty]$, $\beta$, $Z$}

\Indp

$\textstyle \tilde{\tau}_k \sim \beta(\cdot),\quad k=1,\ldots,K$
\tcp*[r]{Sample fractions for action selection}

$\textstyle
a^* \leftarrow \operatorname*{arg\,max}_{a'\in\mathcal A}
\frac{1}{K}
\sum_{k=1}^{K}
Z_{\tilde{\tau}_k}(s',a')$
\tcp*[r]{Select greedy next action}

$\textstyle
\tau_i,\tau'_j \sim \mathcal U([0,1]),
\quad
i=1,\ldots,N,\; j=1,\ldots,N'$
\tcp*[r]{Sample quantile fractions}

$\textstyle
\Delta_i \leftarrow \Delta_p(\tau_i;\epsilon_t)$
\tcp*[r]{Compute robust local correction}

$\textstyle
\widetilde{\delta}_{ij}
\leftarrow
r+\gamma Z_{\tau'_j}(s', a^*)
+\Delta_i
-
Z_{\tau_i}(s,a),
\quad \forall i,j$
\tcp*[r]{Compute robust TD error}

\Indm

\KwOut{$\textstyle
\frac{1}{N'}
\sum_{i=1}^{N}
\sum_{j=1}^{N'}
\rho_{\tau_i}
\left(
\widetilde{\delta}_{ij}
\right)$}

\end{algorithm}




\paragraph{Check loss versus Huberized loss.}
Recent WDRQR theory shows that, for \(p>1\), the check loss is
essentially the unique convex loss that admits an exact additive Wasserstein
regularization under a location-adjusted objective. The quantile Huber loss,
although standard in quantile-based methods like QR-DQN and IQN, is not an affine transformation
of the check loss because it is locally quadratic and globally linear.
Therefore, the exact distributionally robust quantile regression equivalence is not feasible to extend to the
Huber variant. Instead, RQIQN-Huber is used as a practical smooth surrogate
that preserves the same quantile-dependent perturbation structure.

While Corollary~\ref{cor:RQIQNloss} characterizes RQIQN under the original check loss, our implementation adopts a Huberized quantile regression objective for improved optimization stability. Specifically, given $N$ current quantile fractions and $N'$ target quantile fractions, the RQIQN-Huber loss is defined as
\[
    \mathcal L_{\mathrm{RQIQN\text{-}Huber}}
    =
    \frac{1}{N'}
    \sum_{i=1}^{N}
    \sum_{j=1}^{N'}
    \rho_{\tau_i}^{\kappa}
    \left(
    \widetilde\delta_{ij}
    \right),
    \quad
    \widetilde{\delta}_{ij}
    =
    r+\gamma Z_{\tau'_j}(s', a^*)
    +\Delta_i
    -
    Z_{\tau_i}(s,a).
\]

\section{Empirical Results}\label{empiricalresults}

In this section, we report preliminary experimental results. RQIQN is implemented on top of the standard IQN architecture. Unless otherwise specified, we use the check loss for theoretical consistency and set more general $p=2$ for the closed-form Wasserstein correction. The initial robustness radius $\epsilon_0=1$ is natural and stable across the considered tasks. For Atari experiments, we use a reverse-logistic decay schedule with midpoint $t_0=3.75\times 10^6$ training steps and sharpness $k=1.2\times 10^{-6}$. For the navigation experiments, which run for $3\times 10^6$ total training steps, we use the same sharpness $k=1.2\times 10^{-6}$ and set the midpoint to $t_0=5.9\times 10^5$.

\subsection{Risk-Sensitive Control: Autonomous Surface Vehicles Navigation}\label{exp:ASV}


IQN has emerged as a powerful paradigm for robotics tasks~\cite{zhang2025perturbation} such as uncertainty-aware autonomous navigation. 
We evaluate RQIQN in a representative safety-critical marine navigation environment~\cite{lin2023robust}, where bounded workspaces, static obstacles, and vortex-induced flow disturbances are simulated according to physically motivated environmental settings~\cite{acheson1990elementary}.
Following the original learning-based evaluation protocol, we deploy the learned policy for unmanned surface vehicle control and compare RQIQN against IQN and DQN~\cite{mnih2015human}.

\begin{figure*}[h]
  \centering
  \includegraphics[width=0.95\textwidth]
  {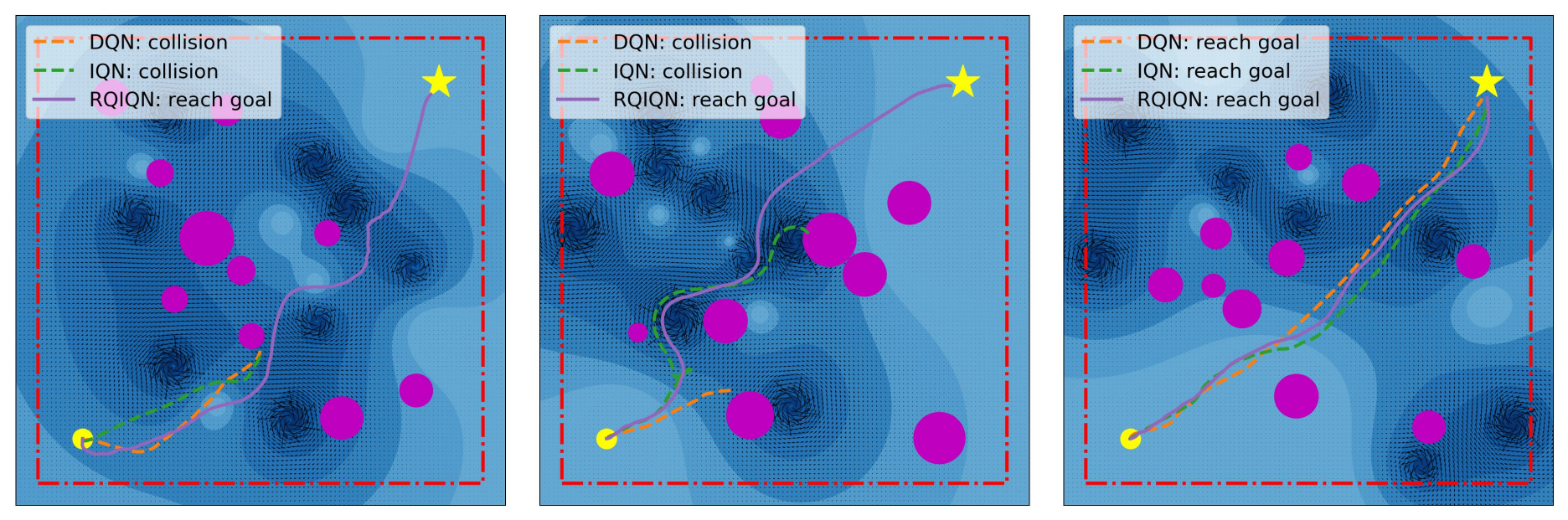}
  \caption{Qualitative trajectory results of RL agents. 
  The \textcolor{yellow!70!orange}{yellow circle} denotes the start position, and the
\textcolor{yellow!70!orange}{yellow star} denotes the goal. 
\textcolor{magenta}{Magenta circles} indicate static obstacles, while the background
vector field represents vortex-induced flow disturbances. Trajectories for IQN and RQIQN are shown under the adaptive setting, where both achieve stronger performance.}
  \label{fig:USVtraj}
\end{figure*}

Specifically, the RL agent receives observations encoding the vessel state and LiDAR-based frontal obstacle distances without environmental prior knowledge, and outputs discrete motion-control actions for goal-directed navigation. The reward encourages goal-reaching progress and safe navigation.
The objective is to learn a policy that reaches the target efficiently while avoiding collisions and maintaining robust behavior under invisible flow-induced disturbances. 

\begin{table}[h]
\centering
\caption{
Evaluation under the \emph{hard-mode} navigation setting with 10 static obstacles and 8 vortex cores. Results are reported as mean $\pm$ standard deviation over 5 independently trained seeds, with each seed evaluated on 500 episodes using different environment layouts.
}
\label{tab:usv_hard_eval}
\resizebox{\linewidth}{!}{
\begin{tabular}{lcccccc}
\toprule
Method
& \makecell{Success(\%)\\$\uparrow$}
& \makecell{Collision(\%)\\$\downarrow$}
& \makecell{Timeout(\%)\\$\downarrow$}
& \makecell{Return\\$\uparrow$}
& \makecell{Time$_{\mathrm{succ}}$(s)\\$\downarrow$}
& \makecell{Energy$_{\mathrm{succ}}$\\$\downarrow$} \\
\midrule
DQN
& 67.24 $\pm$ 4.79
& 32.72 $\pm$ 4.77
& \textbf{0.04 $\pm$ 0.09}
& 12.97 $\pm$ 3.65
& \textbf{33.48 $\pm$ 0.33}
& \textbf{98.95 $\pm$ 5.76} \\

IQN
& 64.80 $\pm$ 22.66
& 21.92 $\pm$ 9.54
& 13.28 $\pm$ 29.47
& -13.17 $\pm$ 51.78
& 39.08 $\pm$ 1.22
& 122.56 $\pm$ 17.29 \\

\rowcolor{gray!12}
\quad +Adaptive
& 71.04 $\pm$ 23.35
& 15.44 $\pm$ 6.13
& 13.52 $\pm$ 29.11
& -12.19 $\pm$ 51.20
& 42.24 $\pm$ 1.55
& 133.31 $\pm$ 17.04 \\

RQIQN
& 83.24 $\pm$ 1.19
& 16.48 $\pm$ 1.14
& 0.28 $\pm$ 0.27
& \textbf{19.20 $\pm$ 1.73}
& 37.55 $\pm$ 1.25
& 111.24 $\pm$ 6.34 \\

\rowcolor{gray!12}
\quad +Adaptive
& \textbf{85.64 $\pm$ 2.27}
& \textbf{14.20 $\pm$ 2.02}
& 0.16 $\pm$ 0.26
& 18.57 $\pm$ 3.90
& 39.19 $\pm$ 1.87
& 115.83 $\pm$ 6.41 \\

\bottomrule
\end{tabular}
}
\end{table}


In this study, the default IQN and RQIQN agents use the natural full quantile support, i.e., $\tau\sim U([0,1])$. Following the existing navigation task, we additionally evaluate the \emph{\textcolor{gray}{adaptive variants}} in Table~\ref{tab:usv_hard_eval}, where the CVaR threshold $\eta$ is adjusted according to the perceived distance to the nearest obstacle. Smaller obstacle distances induce more conservative lower-tail sampling and RQIQN under both sampling reaches improved navigation results and safety. 
As shown in Figure~\ref{fig:USVtraj}, RQIQN remains robust in dense-obstacle scenarios where vortex-induced flows perturb actions and state transitions.



\subsection{Atari Games}\label{exp:Atari}



We compare our algorithm against several representative quantile-based DistRL baselines that have demonstrated strong performance on standard RL benchmarks. DQN serves as the expectation-based baseline that learns only scalar action values. C51~\cite{bellemare2017distributional} represents the return distribution with a categorical distribution over fixed supports. QR-DQN\cite{dabney2018distributional} learns a fixed set of quantile locations via quantile regression, while IQN extends this idea by sampling quantile fractions to approximate an implicit quantile function. Rainbow~\cite{hessel2018rainbow} is included as a strong aggregate baseline that combines C51 with several orthogonal improvements to DQN. RQIQN is evaluated as our robust extension of IQN, using the implicit quantile representation with proposed Wasserstein-robust local correction.

\begin{figure*}[h]
  \centering
  \includegraphics[width=0.95\textwidth]
  {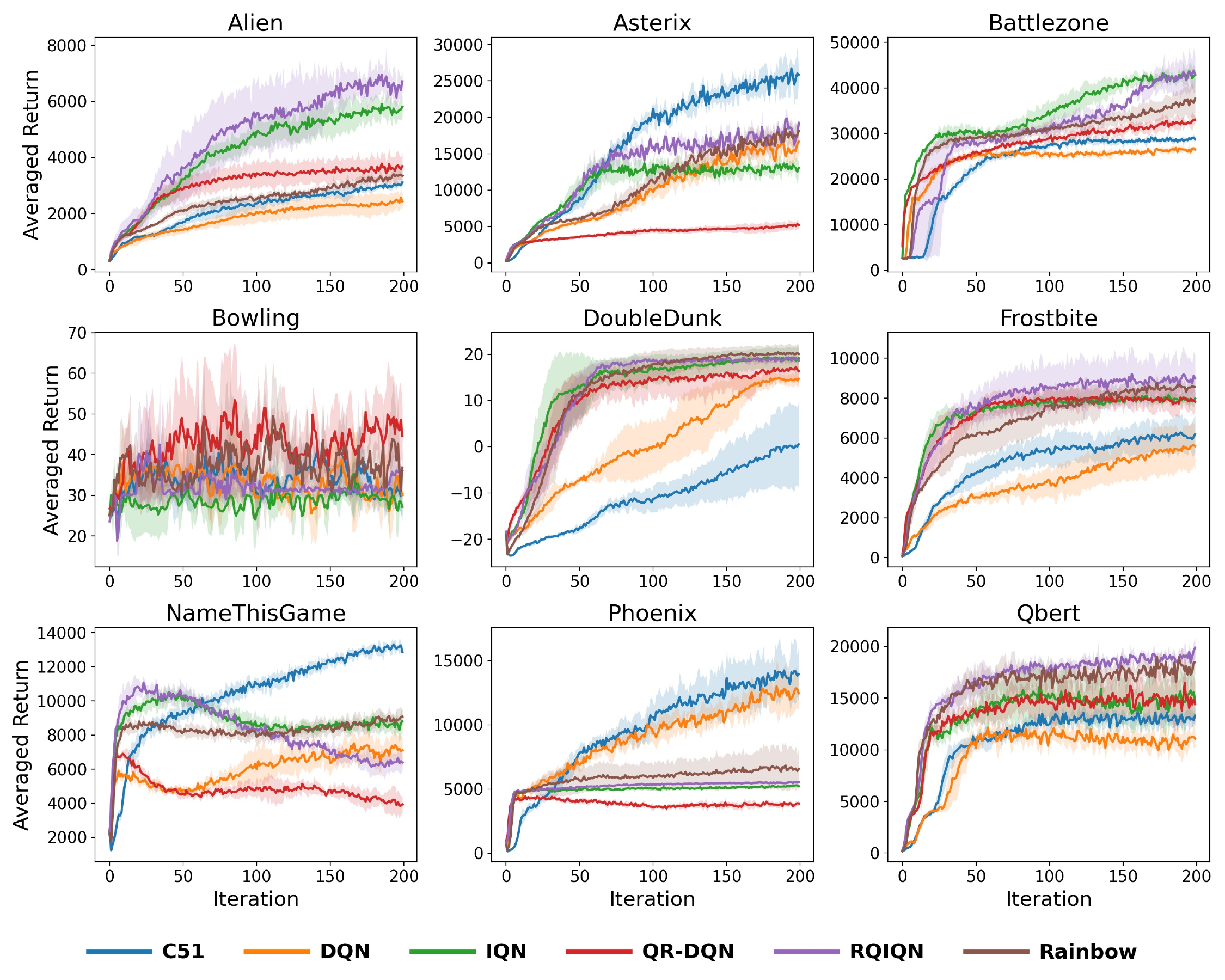}
  \caption{Performance comparison on 9 Atari games. 
  RQIQN results are averaged over 3 random seeds and compared against reference values from Castro et al.~\cite{castro2018dopamine}.}
  \label{fig:atari_curves}
\end{figure*}

We evaluate on several Atari games, including the Atari-5 subbenchmark~\citep{aitchison2023atari}, a compact subset of the Arcade Learning Environment (ALE) designed to approximate trends on the full Atari-57 suite with representativeness and substantially lower computational cost.
For reproducible evaluation, all baselines and our method are implemented in the Dopamine framework~\cite{castro2018dopamine} and trained for 200M frames under the \emph{sticky-actions protocol}~\cite{machado2018revisiting}. Under this protocol, the ALE repeats the previously executed action with probability $0.25$ instead of always applying the agent's newly selected action, injecting controlled stochasticity and reducing reliance on deterministic Atari dynamics.

As shown in Figure~\ref{fig:atari_curves}, RQIQN improves over IQN on most Atari games, with the exception of \textsc{NameThisGame}. Notably, it achieves stronger performance on \textsc{Alien}, \textsc{BattleZone}, \textsc{Frostbite}, and \textsc{Qbert} compared with the evaluated baselines. 
These results suggest that the proposed robust quantile correction can further enhance IQN, indicating that quantile-level robustness remains beneficial even when the quantile Huber loss already provides a stronger practical surrogate than the pure check loss.



\begin{figure*}[h]
  \centering
  \includegraphics[width=0.6\textwidth]
  {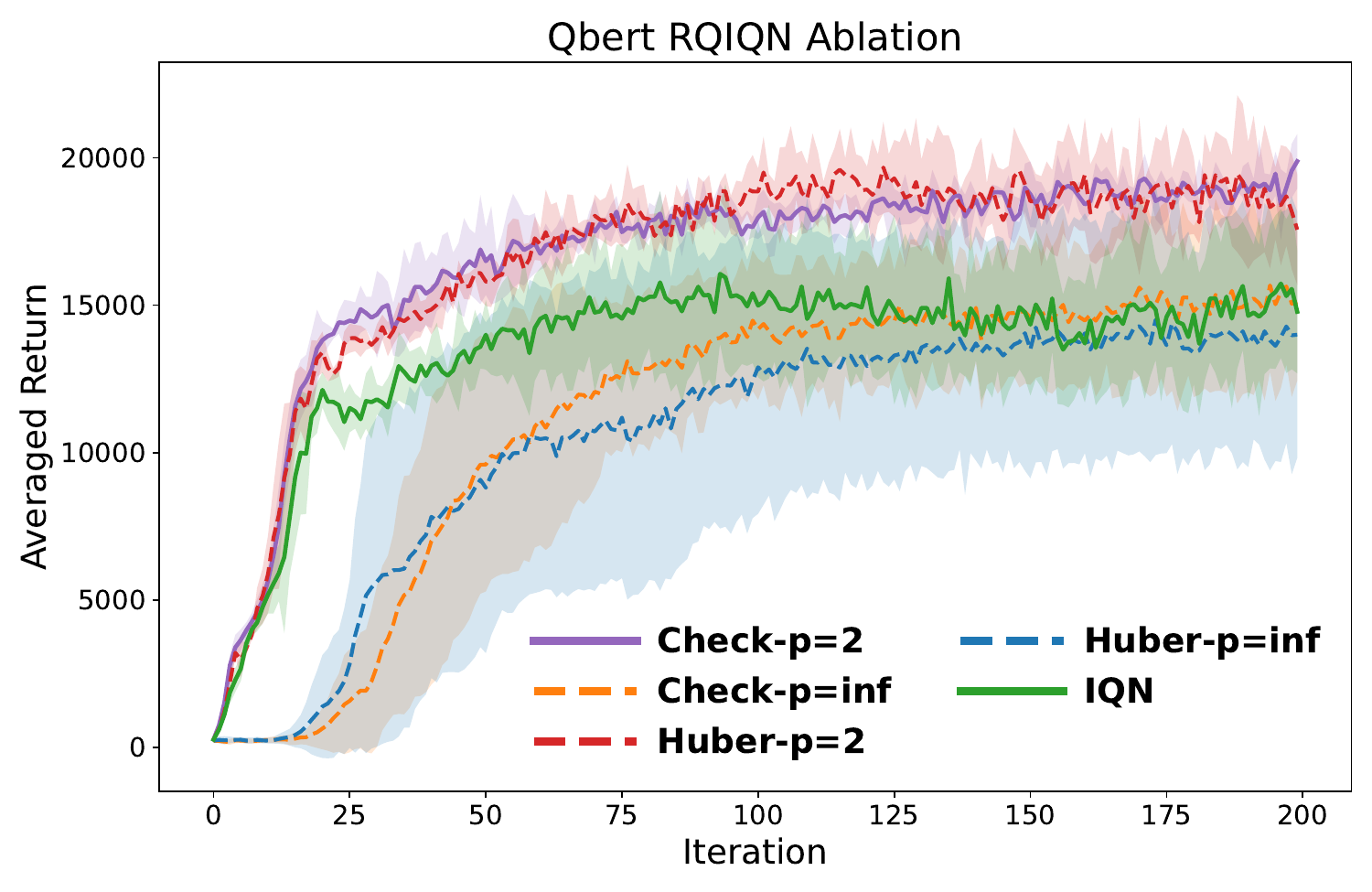}
  \caption{Performance of RQIQN variants.}
  \label{fig:qbert_ablation}
\end{figure*}

\paragraph{Variants of RQIQN.}
To examine the sensitivity of RQIQN to key implementation choices, we further conduct ablation studies on \textsc{Qbert}. Specifically, we compare the check-loss and Huberized variants, and evaluate two representative type-$p$ Wasserstein ambiguity sets, corresponding to $p=2$ and $p=\infty$. 
As shown in Figure~\ref{fig:qbert_ablation}, RQIQN achieves the best performance under the default configuration, using the check loss with a type-$2$ Wasserstein ambiguity set. The Huberized variant with $p=2$ performs comparably, with only a small performance gap, and both variants consistently outperform IQN. In contrast, when the ambiguity set is defined with $p=\infty$, both the check-loss and Huberized variants exhibit performance degradation. These results suggest that RQIQN is more sensitive to the choice of Wasserstein geometry than to the choice between check and Huberized losses.

\paragraph{The Suitability of Distribution Representation.} 
In Figure~\ref{fig:atari_curves}, we observe that on \textsc{NameThisGame}, methods with more advanced schemes exhibit unexpected performance fluctuations and converge to suboptimal final returns. In contrast, C51 maintains a relatively stable and high training return throughout learning.
This indicates that RQIQN, despite its overall gains, may still be vulnerable to performance degradation in specific environments, as observed for other advanced DistRL methods.

In the standard Atari setting, rewards are
clipped to $[-1,1]$, and C51 represents the return distribution using 51 fixed
atoms on a bounded support, typically $[V_{\min},V_{\max}]=[-10,10]$. Bellman
targets are projected back onto this fixed support, which can suppress the
influence of extreme high-return estimates and induce a conservative bias in
action selection. This may prevent the agent from over-committing to aggressive
strategies driven by unstable tail values.

In contrast, IQN and RQIQN use a more flexible quantile-function representation
without a fixed finite support. While this flexibility improves distributional
expressiveness, it also makes the estimated mean return more sensitive to
fluctuations in the distributional tails when the target distribution is not
well fitted. The observation is informative: in principle, a sufficiently accurate
quantile model should exploit this flexibility without suffering from tail
instability, but in practice there is a trade-off between representational
capacity and estimation error. This trade-off is consistent with the motivation
of IQN and RQIQN. Rather than imposing a fixed support, one can exploit the
flexible quantile representation through distortion-based risk control, such as
CVaR sampling, to emphasize different regions of the return distribution. This
also aligns with our navigation results, where adaptive CVaR distortion improves
risk-aware control under hazardous conditions.


\section{Conclusion}
In this work, we introduce RQIQN, which improves IQN by making local quantile fitting robust to finite-sample perturbations of the bootstrapped Bellman target distribution, without introducing computationally intensive schemes. We further develop a Wasserstein-robust quantile correction mechanism for mitigating degeneration problem. 
This correction modulates the distribution geometry during Bellman updates, leading to improved performance on Atari games and risky navigation.

\newpage
\bibliography{main}

@article{mnih2013playing,
  title={Playing atari with deep reinforcement learning},
  author={Mnih, Volodymyr and Kavukcuoglu, Koray and Silver, David and Graves, Alex and Antonoglou, Ioannis and Wierstra, Daan and Riedmiller, Martin},
  journal={arXiv preprint arXiv:1312.5602},
  year={2013}
}

@inproceedings{bellemare2017distributional,
  title={A distributional perspective on reinforcement learning},
  author={Bellemare, Marc G and Dabney, Will and Munos, R{\'e}mi},
  booktitle={International conference on machine learning},
  pages={449--458},
  year={2017},
  organization={Pmlr}
}

@inproceedings{dabney2018implicit,
  title={Implicit quantile networks for distributional reinforcement learning},
  author={Dabney, Will and Ostrovski, Georg and Silver, David and Munos, R{\'e}mi},
  booktitle={International conference on machine learning},
  pages={1096--1105},
  year={2018},
  organization={PMLR}
}

@inproceedings{dabney2018distributional,
  title={Distributional reinforcement learning with quantile regression},
  author={Dabney, Will and Rowland, Mark and Bellemare, Marc and Munos, R{\'e}mi},
  booktitle={Proceedings of the AAAI conference on artificial intelligence},
  volume={32},
  number={1},
  year={2018}
}

@article{jullien2023distributional,
  title={Distributional reinforcement learning with dual expectile-quantile regression},
  author={Jullien, Sami and Deffayet, Romain and Renders, Jean-Michel and Groth, Paul and de Rijke, Maarten},
  journal={arXiv preprint arXiv:2305.16877},
  year={2023}
}

@inproceedings{rowland2019statistics,
  title={Statistics and samples in distributional reinforcement learning},
  author={Rowland, Mark and Dadashi, Robert and Kumar, Saurabh and Munos, R{\'e}mi and Bellemare, Marc G and Dabney, Will},
  booktitle={International Conference on Machine Learning},
  pages={5528--5536},
  year={2019},
  organization={PMLR}
}

@article{yang2019fully,
  title={Fully parameterized quantile function for distributional reinforcement learning},
  author={Yang, Derek and Zhao, Li and Lin, Zichuan and Qin, Tao and Bian, Jiang and Liu, Tie-Yan},
  journal={Advances in neural information processing systems},
  volume={32},
  year={2019}
}

@article{zhang2026wasserstein,
  title={Wasserstein Distributionally Robust Quantile Regression},
  author={Zhang, Chunxu and Mao, Tiantian and Wang, Ruodu},
  journal={arXiv preprint arXiv:2603.14991},
  year={2026}
}

@article{zhang2025perturbation,
  title={Perturbation-mitigated USV Navigation with Distributionally Robust Reinforcement Learning},
  author={Zhang, Zhaofan and Yang, Minghao and Xie, Sihong and Xiong, Hui},
  journal={arXiv preprint arXiv:2512.00030},
  year={2025}
}

@inproceedings{lin2023robust,
  title={Robust unmanned surface vehicle navigation with distributional reinforcement learning},
  author={Lin, Xi and McConnell, John and Englot, Brendan},
  booktitle={2023 IEEE/RSJ International Conference on Intelligent Robots and Systems (IROS)},
  pages={6185--6191},
  year={2023},
  organization={IEEE}
}

@inproceedings{shi2024robust,
  title={Robust quadrupedal locomotion via risk-averse policy learning},
  author={Shi, Jiyuan and Bai, Chenjia and He, Haoran and Han, Lei and Wang, Dong and Zhao, Bin and Zhao, Mingguo and Li, Xiu and Li, Xuelong},
  booktitle={2024 IEEE International Conference on Robotics and Automation (ICRA)},
  pages={11459--11466},
  year={2024},
  organization={IEEE}
}

@incollection{huber1992robust,
  title={Robust estimation of a location parameter},
  author={Huber, Peter J},
  booktitle={Breakthroughs in statistics: Methodology and distribution},
  pages={492--518},
  year={1992},
  publisher={Springer}
}

@inproceedings{zhang2024distributional,
  title={Distributional Reinforcement Learning with Sample-set Bellman Update},
  author={Zhang, Weijian and Wang, Jianshu and Yu, Yang},
  booktitle={2024 IEEE International Conference on Robotics and Automation (ICRA)},
  pages={2852--2858},
  year={2024},
  organization={IEEE}
}

@article{nguyen2020distributional,
  title={Distributional reinforcement learning with maximum mean discrepancy},
  author={Nguyen, Thanh Tang and Gupta, Sunil and Venkatesh, Svetha},
  journal={Association for the Advancement of Artificial Intelligence (AAAI)},
  year={2020}
}

@article{newey1987asymmetric,
  title={Asymmetric least squares estimation and testing},
  author={Newey, Whitney K and Powell, James L},
  journal={Econometrica: Journal of the Econometric Society},
  pages={819--847},
  year={1987},
  publisher={JSTOR}
}

@article{dempster1977maximum,
  title={Maximum likelihood from incomplete data via the EM algorithm},
  author={Dempster, Arthur P and Laird, Nan M and Rubin, Donald B},
  journal={Journal of the royal statistical society: series B (methodological)},
  volume={39},
  number={1},
  pages={1--22},
  year={1977},
  publisher={Wiley Online Library}
}

@article{tzavidis2010robust,
  title={Robust estimation of small-area means and quantiles},
  author={Tzavidis, Nikos and Marchetti, Stefano and Chambers, Ray},
  journal={Australian \& New Zealand Journal of Statistics},
  volume={52},
  number={2},
  pages={167--186},
  year={2010},
  publisher={Wiley Online Library}
}

@article{john2015robustness,
  title={Robustness of quantile regression to outliers},
  author={John, Onyedikachi O},
  journal={American Journal of Applied Mathematics and Statistics},
  volume={3},
  number={2},
  pages={86--88},
  year={2015}
}

@article{galarza2017robust,
  title={Robust quantile regression using a generalized class of skewed distributions},
  author={Galarza Morales, Christian and Lachos Davila, Victor and Barbosa Cabral, Celso and Castro Cepero, Luis},
  journal={Stat},
  volume={6},
  number={1},
  pages={113--130},
  year={2017},
  publisher={Wiley Online Library}
}

@inproceedings{aitchison2023atari,
  title={Atari-5: Distilling the arcade learning environment down to five games},
  author={Aitchison, Matthew and Sweetser, Penny and Hutter, Marcus},
  booktitle={International Conference on Machine Learning},
  pages={421--438},
  year={2023},
  organization={PMLR}
}

@article{mnih2015human,
  title={Human-level control through deep reinforcement learning},
  author={Mnih, Volodymyr and Kavukcuoglu, Koray and Silver, David and Rusu, Andrei A and Veness, Joel and Bellemare, Marc G and Graves, Alex and Riedmiller, Martin and Fidjeland, Andreas K and Ostrovski, Georg and others},
  journal={nature},
  volume={518},
  number={7540},
  pages={529--533},
  year={2015},
  publisher={Nature Publishing Group}
}

@inproceedings{hessel2018rainbow,
  title={Rainbow: Combining improvements in deep reinforcement learning},
  author={Hessel, Matteo and Modayil, Joseph and Van Hasselt, Hado and Schaul, Tom and Ostrovski, Georg and Dabney, Will and Horgan, Dan and Piot, Bilal and Azar, Mohammad and Silver, David},
  booktitle={Proceedings of the AAAI conference on artificial intelligence},
  volume={32},
  number={1},
  year={2018}
}

@article{castro2018dopamine,
  title={Dopamine: A research framework for deep reinforcement learning},
  author={Castro, Pablo Samuel and Moitra, Subhodeep and Gelada, Carles and Kumar, Saurabh and Bellemare, Marc G},
  journal={arXiv preprint arXiv:1812.06110},
  year={2018}
}

@article{machado2018revisiting,
  title={Revisiting the arcade learning environment: Evaluation protocols and open problems for general agents},
  author={Machado, Marlos C and Bellemare, Marc G and Talvitie, Erik and Veness, Joel and Hausknecht, Matthew and Bowling, Michael},
  journal={Journal of Artificial Intelligence Research},
  volume={61},
  pages={523--562},
  year={2018}
}

@book{acheson1990elementary,
  title={Elementary fluid dynamics},
  author={Acheson, David J},
  year={1990},
  publisher={Oxford University Press}
}
\bibliographystyle{unsrt}


\end{document}